\documentclass[conference,letterpaper]{IEEEtran}
\IEEEoverridecommandlockouts
\usepackage{cite}
\usepackage{amsmath,amssymb,amsfonts}
\usepackage{algorithmic}
\usepackage{graphicx}
\usepackage{textcomp}
\usepackage{xcolor}
\def\BibTeX{{\rm B\kern-.05em{\sc i\kern-.025em b}\kern-.08em
    T\kern-.1667em\lower.7ex\hbox{E}\kern-.125emX}}
    
\usepackage{url}            
\usepackage{booktabs}       
\usepackage{nicefrac}       
\usepackage{microtype}      
\usepackage{subfigure}
\usepackage{booktabs} 
\usepackage{bm} 
\usepackage{bbm}

\graphicspath{{figs/}}

\begin{document}

\title{Recurrent Neural Filters: Learning Independent Bayesian Filtering Steps for Time Series Prediction
}

\author{\IEEEauthorblockN{Bryan Lim, Stefan Zohren and Stephen Roberts}
\IEEEauthorblockA{\textit{Oxford-Man Institute of Quantitative Finance} \\
\textit{Department of Engineering Science} \\
\textit{University of Oxford}\\
Oxford, UK\\
\{blim,zohren,sjrob\}@robots.ox.ac.uk}
}

\maketitle

\begin{abstract}
Despite the recent popularity of deep generative state space models, few comparisons have been made between network architectures and the inference steps of the Bayesian filtering framework -- with most models simultaneously approximating both state transition and update steps with a single recurrent neural network (RNN). In this paper, we introduce the Recurrent Neural Filter (RNF), a novel recurrent autoencoder architecture that learns distinct representations for each Bayesian filtering step, captured by a series of encoders and decoders. Testing this on three real-world time series datasets, we demonstrate that the decoupled representations learnt improve the accuracy of one-step-ahead forecasts while providing realistic uncertainty estimates, and also facilitate multistep prediction through the separation of encoder stages.
\end{abstract}


\section{Introduction}
Bayesian filtering \cite{BayesianSigProc} has been extensively used within the domain of time series prediction, with numerous applications across different fields -- including target tracking \cite{TrackingTextbook}, robotics \cite{StateEstimationForRobots}, finance \cite{StochVolKalmanFilter}, and medicine \cite{DiseaseProgressionModels}. Performing inference via a series of prediction and update steps \cite{BayesianFilteringAndSmoothing}, Bayesian filters recursively update the posterior distribution of predictions -- or the belief state \cite{ProbabilisticRobotics} -- with the arrival of new data. For many filter models -- such as the Kalman filter \cite{KalmanFilterPaper} and the unscented Kalman filter \cite{UnscentedKalmanFilter} -- deterministic functions are used at each step to adjust the sufficient statistics of the belief state, guided by generative models of the data. Each function quantifies the impact of different sources of information on latent state estimates -- specifically time evolution and exogenous inputs in the prediction step, and realised observations in the update step. On top of efficient inference and uncertainty estimation, this decomposition of inference steps enables Bayes filters to be deployed in use cases beyond basic one-step-ahead prediction -- with simple extensions for multistep prediction \cite{KalmanFilterMultistepExample} and prediction in the presence of missing observations \cite{KalmanFilterMissingExample1}. 

With the increasing use of deep neural networks for time series prediction, applications of recurrent variational autoencoder (RVAE) architectures have been investigated for forecasting non-linear state space models \cite{VRNN,DeepKalmanFilters,StructuredInferenceNetworks,DeepVariationalBayesFilters}. Learning dynamics directly from data, they avoid the need for explicit model specification -- overcoming a key limitation in standard Bayes filters. However, these RVAEs focus predominantly on encapsulating the generative form of the state space model -- implicitly condensing both state transition and update steps into a single representation learnt by the RVAE decoder -- and make it impossible to decouple the Bayes filter steps. 

Recent works in deep generative modelling have focused on the use of neural networks to learn independent factors of variation in static datasets -- through the encouragement of disentangled representations \cite{DisentangledRepresentations,DisentanglingByFactorising} or by learning causal mechanisms \cite{IndependentCausalMechanisms,DisentanglingControllableFactors}. While a wide range of training procedures and loss functions have been proposed \cite{ChallengingDisentangledVAEAssumptions}, methods in general use dedicated network components to learn distinct interpretable relationships  -- ranging from orthogonalising latent representations in variational autoencoders \cite{BetaVAE} to learning independent modules for different causal pathways \cite{IndependentCausalMechanisms}. By understanding the relationships encapsulated by each component, we can subsequently decouple them for use in related tasks -- allowing the learnt mechanisms to generalise to novel domains \cite{IndependentCausalMechanisms,BuidlingMachinesLikePeople} or to provide building blocks for transfer learning \cite{BetaVAE}.

In this paper, we introduce the Recurrent Neural Filter (RNF) -- a novel recurrent autoencoder architecture which aligns network modules (encoders and decoders) with the inference steps of the Bayes filter -- making several contributions over standard approaches. Firstly, we propose a new training procedure to encourage independent representations within each module, by directly training intermediate encoders with a common emission decoder. In doing so, we augment the loss function with additional regularisation terms (see Section \ref{sec:training}), and directly encourage each encoder to learn functions to update the filter's belief state given available information. Furthermore, to encourage the decoupling of encoder stages, we randomly drop out the input dynamics and error correction encoders during training -- which can be viewed as artificially introducing missingness to the inputs and observations respectively. Finally, we highlight performance gains for one-step-ahead predictions through experiments on 3 real-world time series datasets, and investigate multistep predictions as a use case for generalising the RNF's decoupled representations to other tasks -- demonstrating performance improvements from the recursive application of the state transition encoders alone.
\section{Related Work}
\textbf{RVAEs for State Space Modelling: \quad} The work of \cite{VRNN} identifies close parallels between RNNs and latent state space models, both consisting of an internal hidden state that drives output forecasts and observations. Using an RVAE architecture described as a variational RNN (VRNN), they build their recognition network (encoder) with RNNs and produce samples for the stochastic hidden state at each time point. Deep Kalman filters (DKFs) \cite{DeepKalmanFilters,StructuredInferenceNetworks} take this a step further by allowing for exogenous inputs in their network and incorporating a special KL loss term to penalise state transitions between time steps. Deep Variational Bayes Filters (DVBFs) \cite{DeepVariationalBayesFilters} enhance the interpretability of DKFs by modelling state transitions with parametric -- e.g. linear -- models, which take in stochastic samples from the recognition model as inputs. In general, while the above models capture the generative modelling aspects of the state space framework, their inference procedure blends both state transition and error correction steps, obliging the recognition model to learn representations for both simultaneously. In contrast, the RNF uses separate neural network components to directly model the Bayes filter steps -- leading to improvements in representation learning and enhanced predictive performance in time series applications.

\textbf{Hybrid Approaches: \quad} In \cite{JohnsonNIPS}, the authors take a hybrid approach with the structured variational autoencoder (SVAE), proposing an efficient general inference framework that combines probabilistic graphical models for the latent state with neural network observation models. This is similar in spirit to the Kernel Kalman Filter \cite{KernelKalmanFilter}, allowing for predictions to be made on complex observational datasets -- such as raw images -- by encoding high dimensional outputs onto a lower dimensional latent representation modelled with a dynamical systems model. Although SVAEs provide a degree of interpretability to temporal dynamics, they also require a parametric model to be defined for the latent states which may be challenging for arbitrary time series datasets. The RNF, in comparison, can learn the relationships directly from data, without the need for explicit model specification. The Kalman variational autoencoder (KVAE) \cite{KVAE} extends ideas from the SVAE, modelling latent state using a linear Gaussian state space model (LGSSM). To allow for non-linear dynamics, the KVAE uses a recognition model to produce time-varying parameters for the LGSSM, weighting a set of $K$ constant parameters using weights generated by a neural network. Deep State Space Models (DSSM) \cite{DeepStateSpaceModels} investigate a similar approach within the context of time series prediction, using an RNN to generate parameters of the LGSSM at each time step. While the LGSSM components do allow for the application of the Kalman filter, we note that updates to the time-varying weights from the RNN once again blend the prediction and update steps -- making the separation of Bayes filter steps and generalisation to other tasks non-trivial. On the other hand, the RNF naturally supports simple extensions (e.g. multistep prediction) similarly to other Bayes filter -- due to the close alignment of the RNF architecture with the Bayes filter steps and the use of decoupled representations across encoders and decoders.

\textbf{Autoregressive Architectures: \quad} An alternative approach to deep generative modelling focuses on the autoregressive factorisation of the joint distribution of observations  $\left(\text{i.e. }p(\bm{y}_{1:T}) = \prod_t  p(\bm{y}_{t} | \bm{y}_{1:t})\right)$, directly generating the conditional distribution at each step. For instance, WaveNet \cite{WaveNet} and Transformer \cite{transformer,transformerXL} networks use dilated CNNs and attention-based models to build predictive distributions. While successful in speech generation and language applications, these models suffer from several limitations in the context of time series prediction. Firstly, the CNN and attention models require the pre-specification of the amount of relevant history to use in predictions -- with the size of the look-back window controlled by the length of the receptive field or extended context -- which may be difficult when the data generating process is unknown. Furthermore, they also rely on a discretisation of the output, generating probabilities of occurrence within each discrete interval using a softmax layer. This can create generalisation issues for time series where outputs are unbounded. In contrast, the LSTM cells used in the RNF recognition model remove the need to define a look-back window, and the parametric distributions used for outputs are compatible with unbounded continuous observations.

In other works, the use of RNNs in autoregressive architectures for time series prediction have been explored in DeepAR models \cite{DeepAR}, where LSTM networks output Gaussian mean and standard deviation parameters of predictive distributions at each step. We include this as a benchmark in our tests, noting the improvements observed with the RNF through its alignment with the Bayesian filtering paradigm.

\textbf{Predictive State Representations: \quad} Predictive state RNNs (PSRNN) \cite{OPSRNN,PSRNN,PSDecoders} use an alternative formulation of the Bayes filter, utilising a state representation that corresponds to the statistics of the predictive distribution of future observations. Predictions are made using a two-stage regression approach modelled by their proposed architectures. Compared to alternative approaches, PSRNNs only produce point estimates for their forecasts -- lacking the uncertainty bounds from predictive distributions produced by the RNF. 

\textbf{Non-Parametric State Space Models: \quad} Gaussian Process state space models (GP-SSMs)  \cite{StateSpaceGPs,NonGaussianStateSpaceGPs}  and variational approximations \cite{GPSSRNN}, provide an alternative non-parametric approach to forecasting non-linear state space models -- modelling hidden states and observation dynamics using GPs. While they have similar benefits to Bayes filters (i.e. predictive uncertainties, natural multistep prediction etc.), inference at each time step has at least an $O(T)$ complexity in the number of past observations -- either via sparse GP approximations or Kalman filter formulations \cite{GPComparisonsToKF}. In contrast, the RNF updates its belief state at each time point only with the latest observations and input, making it suitable for real-time prediction on high-frequency datasets.

\textbf{RNNs for Multistep Prediction: \quad} Customised sequence-to-sequence architectures have been explored in \cite{Modred,MultistepQuantileRecurrentForecaster} for multistep time series prediction, typically predefining the forecast horizon, and using computationally expensive customised training procedures to improve performance. In contrast, the RNF does not require the use of a separate training procedure for multistep predictions -- hence reducing the computational overhead -- and does not require the specification of a fixed forecast horizon. 

\section{Problem Definition}
\begin{figure*}[tbh]
\centerline{\includegraphics[width=1.0\linewidth]{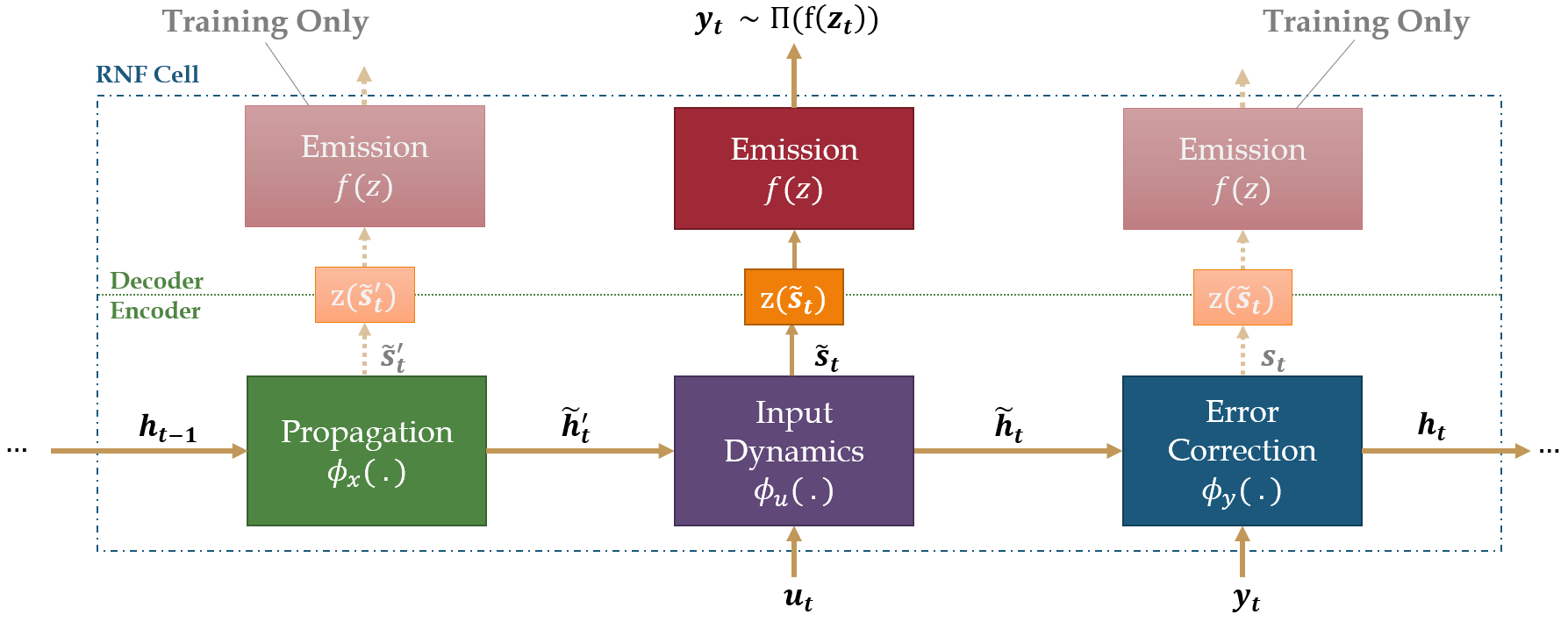}}
\caption{RNF Network Architecture}
\label{fig:architecture}
\end{figure*} 	
Let $\bm{y}_t = \left[ y_t(1), \dots, y_t(O) \right]^T$ be a vector of observations, driven by a set of stochastic hidden states $\bm{x}_t = [ x_t(1), \dots, x_t(J) ]^T$ and exogenous inputs $\bm{u}_t = \left[ u_t(1), \dots, u_t(I) \right]^T$ . We consider non-linear state space models of the following form:
\begin{align}
\label{eqn:generative_model_y}
\bm{y}_t &\sim \Pi\big(~f(\bm{x}_t~) \big)\\
\bm{x}_t &\sim N \big(~\mu(\bm{x}_{t-1}, \bm{u}_t),~\Sigma(\bm{x}_{t-1}, \bm{u}_t) ~\big)
\label{eqn:generative_model_x}
\end{align}

where $\Pi$ is an arbitrary distribution parametrised by a non-linear function $f(\bm{x}_t)$, with $\mu(\cdot)$ and $\Sigma(\cdot)$ being mean and covariance functions respectively.

Bayes filters allow for efficient inference through the use of a belief state, i.e. a posterior distribution of hidden states given past observations $\bm{y}_{1:t} =  \{\bm{y}_1 , \dots, \bm{y}_t \}$ and inputs $\bm{u}_{1:t} =  \{\bm{u}_1 , \dots, \bm{u}_t \}$. This is achieved through the maintenance of a set sufficient statistics $\bm{\theta}_t$ -- e.g. means and covariances $\bm{\theta}_t \in \{ \bm{\mu}_t, \bm{\Sigma}_t \}$ -- which compactly summarise the historical data: 
\begin{align}
p(\bm{x}_t | \bm{y}_{1:t}, \bm{u}_{1:t}) &= bel(\bm{x}_t ; \bm{\theta}_t) \\
& = N\left(\bm{x}_t ;~ \bm{\mu}_t, \bm{\Sigma}_t \right)
\label{eqn:belief_state}
\end{align}
where $bel(.)$ is a probability distribution function for the belief state. 

For filters such as the Kalman filter -- and non-linear variants like the unscented Kalman filter \cite{UnscentedKalmanFilter} -- $\bm{\theta}_t$ is recursively updated through a series of prediction and update steps which take the general form:\\

\noindent \textbf{Prediction (State Transition):}
\begin{align}
\tilde{\bm{\theta}}_t = \phi_u \left(\bm{\theta}_{t-1}, \bm{u}_t\right) 
\label{eqn:filter_state_transition}
\end{align}

\noindent\textbf{Update (Error Correction):}
\begin{align}
\bm{\theta}_t = \phi_y \left(\tilde{\bm{\theta}}_{t}, \bm{y}_t\right) 
\label{eqn:filter_state_update}
\end{align}
where $\phi_u(\cdot)$ and $\phi_y(\cdot)$ are non-linear deterministic functions.
Forecasts can then be computed using one-step ahead predictive distributions:
\begin{equation}
\label{eqn:vae_filter_form}
p(\bm{y}_t | \bm{y}_{1:t-1}, \bm{u}_{1:t})  = \int p(\bm{y}_t | \bm{x}_t) ~bel\left(\bm{x}_t ; \tilde{\bm{\theta}}_t \right)~ d\bm{x}_t.
\end{equation}
In certain cases -- e.g. with the Kalman filter -- the predictive distribution can also be directly parameterised using analytical functions $g(.)$ for belief state statistics :
\begin{align}
\label{eqn:ae_filter_form}
p(\bm{y}_t | \bm{y}_{1:t-1}, \bm{u}_{1:t})  = p\left(\bm{y}_t |~g(\tilde{\bm{\theta}}_t) \right).
\end{align}
When observations are continuous, such as in standard linear Gaussian state space models, $\bm{y}_t$ can be modelled using a Normal distribution -- i.e. $\bm{y}_t \sim N\left(g_\mu(\tilde{\bm{\theta}}_t),~g_\Sigma(\tilde{\bm{\theta}}_t)\right)$.

\section{Recurrent Neural Filter}
\label{sec:rnf_section}
Recurrent Neural Filters use a series of encoders and decoders to learn independent representations for the Bayesian filtering steps. We investigate two RNF variants as described below, based on Equations \eqref{eqn:vae_filter_form} and \eqref{eqn:ae_filter_form} respectively.

\textbf{Variational Autoencoder Form (VRNF) \quad} Firstly, we capture the belief state of Equation \eqref{eqn:belief_state} using a recurrent VAE-based architecture. At run time, samples of $\bm{x}_t$ are generated from the encoder -- approximating the integral of Equation \eqref{eqn:vae_filter_form} to compute the predictive distribution of $\bm{y}_t$.

\textbf{Standard Autoencoder Form (RNF)\footnote{An open-source implementation of the standard RNF can be found at: {\color{blue}\underline{\url{https://github.com/sjblim/rnf-ijcnn-2020}}}} \quad} Much recent work has demonstrated the sensitivity of VAE performance to the choice of prior distribution, with suboptimal priors either having an ``over-regularising" effect on the loss function during training \cite{ImplicitOptimalPriors,VampPrior,KLAnnealingOld}, or leading to posterior collapse \cite{PosteriorCollapseDef}. As such, we also implement an autoregressive version of the RNF based on Equation \eqref{eqn:ae_filter_form} -- directly feeding encoder latent states into the common emission decoder.

A general architecture diagram for both forms is shown in Figure \ref{fig:architecture}, with the main differences encapsulated within $z(s)$ (see Section \ref{sec:architecture}). 

\subsection{Network Architecture}
\label{sec:architecture}

First, let $\bm{s}_t$ be a latent state that maps to sufficient statistics $\bm{\theta}_t$, which are obtained as outputs from our recognition model. Per Equations \eqref{eqn:filter_state_transition} and \eqref{eqn:filter_state_update}, inference at run-time is controlled through the recursive update of $\bm{s}_t$, using a series of Long Short-Term Memory (LSTM) \cite{LSTM} encoders with exponential linear unit (ELU) activations \cite{ELU}.

\textbf{Encoder \quad} To directly estimate the impact of exogenous inputs on the belief state, the prediction step, Equation \eqref{eqn:filter_state_transition}, is divided into two parts with separate LSTM units $\phi_x(\cdot)$ and $\phi_u(\cdot)$. We use $\bm{h}_t$ to represent all required memory components -- i.e. both output vector and cell state for the standard LSTM -- with $\bm{s}_t$ being the output of the cell. A third LSTM cell $\phi_y(\cdot)$ is then used for the update step, Equation \eqref{eqn:filter_state_update}, with the full set of equations below.\\

\noindent \textbf{Prediction:}
\begin{align}
\label{eqn:encoder_prop}
&\text{\small\textit{Propagation}} && \left[\bm{\tilde{s}}^{'}_t, \bm{\tilde{h}}^{'}_t \right] = \phi_x(\bm{h}_{t-1}) & \\ 
&\text{\small\textit{Input Dynamics}} && \left[ \bm{\tilde{s}}_t, \bm{\tilde{h}}_t \right] = \phi_u(\bm{\tilde{h}}^{'}_t, \bm{u}_t) &
\label{eqn:encoder_input}
\end{align}

\noindent \textbf{Update:}
\begin{align}
&\text{\textit{\small Error Correction}} && \left[ \bm{s}_t, \bm{h}_t \right] = \phi_y(\bm{\tilde{h}}_t, \bm{y}_t) &
\label{eqn:encoder_correction}
\end{align}

For the variational RNF, hidden state variable $\bm{x}_t$ is modelled as multivariate Gaussian, given by: 
\begin{align}
\bm{x}_t &\sim N (m(\tilde{\bm{s}}_t), V(\tilde{\bm{s}}_t))\\
m(\tilde{\bm{s}}_t) &= \bm{W}_m \tilde{\bm{s}}_t + \bm{b}_m\\
\label{eqn:vrnf_encoder_mean}
V(\tilde{\bm{s}}_t) &= \text{diag}(\sigma(\tilde{\bm{s}}_t)\odot\sigma(\tilde{\bm{s}}_t)) \\
\sigma(\tilde{\bm{s}}_t) &=  \text{Softplus}(\bm{W}_\sigma \tilde{\bm{s}}_t + \bm{b}_\sigma),
\label{eqn:vrnf_encoder_std}
\end{align}
where $\bm{W}_{(\cdot)}, \bm{b}_{(\cdot)}$ are the weights/biases of each layer, and $\odot$ is an element-wise (Hadamard) product. \\

For the standard RNF, the encoder state $\tilde{\bm{s}}_t$ is directly fed into the emission decoder leading to the following forms for $ \tilde{\bm{z}}_t = z(\tilde{\bm{s}}_t)$:
\begin{align}
z_{\text{VRNF}} (\tilde{\bm{s}}_t) = \bm{x}_t, &&
z_{\text{RNF}} (\tilde{\bm{s}}_t) = \tilde{\bm{s}}_t.
\end{align}
While the connection to non-linear state-space models facilitates our interpretation of $z_{\text{RNF}} (\tilde{\bm{s}}_t)$, we note that the standard RNF no longer relies on an explicit generative model for the latent state $\bm{x}_t$. This potentially allows the standard RNF to learn more complex update rules for non-Gaussian latent states.

\textbf{Decoder \quad} Given an encoder output $\tilde{\bm{z}}_t$, we use a multi-layer perceptron to model the emission function $f(\cdot)$:
\begin{align}
f (\tilde{\bm{z}}_t)  = \bm{W}_{z_2}~\text{ELU}(\bm{W}_{z_1} ~ \tilde{\bm{z}}_t + \bm{b}_{z_1} )   + \bm{b}_{z_2}.
\end{align}

This allows us to handle both continuous or binary observations using the output models below:
\begin{equation}
\bm{y}_t ^{~\text{continuous}}  \sim N \bigg(f_\mu(\tilde{\bm{z}}_t)~, \Gamma(\tilde{\bm{z}}_t) \bigg), 
\label{eqn:continuous_observation_model}
\end{equation}
\begin{equation}
\bm{y}_t ^{~\text{binary}}  \sim \text{Bernoulli}\bigg( \text{Sigmoid} ( f(\tilde{\bm{z}}_t) ) \bigg). 
\label{eqn:binary_observation_model}
\end{equation}
where $\Gamma(\tilde{\bm{z}}_t)  = diag \left( g_\sigma(\tilde{\bm{z}}_t)  \right)$ is a time-dependent diagonal covariance matrix, and $ g_\sigma(\tilde{\bm{z}}_t) = \text{Softplus}( f_\sigma(\tilde{\bm{z}}_t))$.\\

For $\bm{y}_t ^{~\text{continuous}} $, the weights $W_{z_1}$, $b_{z_1}$ are shared between $f_\mu(\cdot)$ and $f_\sigma(\cdot)$ -- i.e. both observation means and covariances are generated from the same encoder hidden layer.

\subsection{Handling Missing Data and Multistep Prediction}
\label{sec:rnf_missing_data}
\begin{figure}[h]
\centerline{\includegraphics[width=\linewidth]{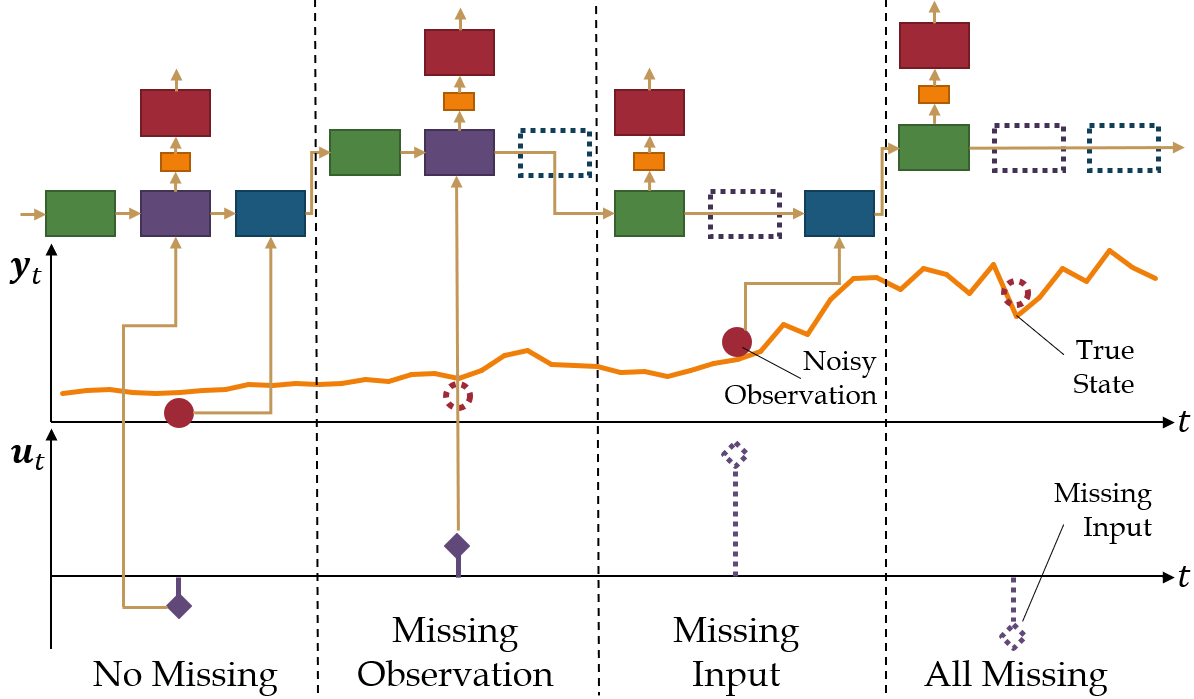}}
\caption{RNF Configuration with Missing Data}
\label{fig:missingness}
\end{figure} 	
%
From the above, we can see that each encoder learns how specific inputs (i.e. time evolution, exogenous inputs and the target) modify the belief state. As such, in a similar fashion to Bayes filters, we decouple the RNF stages at run-time based on the availability of inputs for prediction -- allowing it to handle applications involving missing data or multistep forecasting. 

Figure \ref{fig:missingness} demonstrates how the RNF stages can be combined to accommodate missing data, noting that the colour scheme of the encoders/decoders shown matches that of Figure \ref{fig:architecture}. From the schematic, the propagation encoder -- which is responsible for changes to the belief state due to time evolution -- is always applied, with the input dynamics and error correction encoders only used when inputs or observations are observed respectively. Where inputs are available, the emission decoder is applied to the input dynamics encoder to generate predictions at each step. Failing that, the decoder is applied to the propagation encoder alone.Multistep forecasts can also be treated as predictions in the absence of inputs or observations, with the encoders used to project the belief state in a similar fashion to missing data.
\section{Training Methodology}
\label{sec:training}
Considering the joint probability for a trajectory of length $T$, we train the standard RNF by minimising the negative log-likelihood of the observations. For continuous observations, this involves Gaussian likelihoods from Equation \eqref{eqn:continuous_observation_model}:
\begin{align}
\mathcal{L}_{\text{RNF}} (\bm{\omega}, \tilde{\bm{s}}_{1:T}) &= - \sum_{t=1}^T \log p(\bm{y}_{t} | \tilde{\bm{s}}_t), \\
\log p(\bm{y}_{t} |\tilde{\bm{s}}_t ) &=  -\frac{1}{2} \sum_{j=1}^J \bigg\{  \log(2\pi g_\sigma(j, \tilde{\bm{s}}_t)^2) \nonumber \\
&+ \left\lVert \frac{ \bm{y}_t(j) -  f_\mu \left(j, \tilde{\bm{s}}_t \right)} { g_\sigma(j, \tilde{\bm{s}}_t)} \right\rVert^2 \bigg\} ,
\end{align}
where $\bm{\omega}$ are the weights of the deep neural network, $f_\mu \left(j, \tilde{\bm{z}}_t \right)$ is the $j$-th element of $f_\mu \left( \tilde{\bm{z}}_t \right)$, and $g_\sigma(j, \tilde{\bm{z}}_t)$ the $j$-th element of $g_\sigma(\tilde{\bm{z}}_t)$.

For the VRNF, we adopt the Stochastic Gradient Variational Bayes (SGVB) estimator of \cite{VAE} for our VAE evidence lower bound, expressing our loss function as:
\begin{align}
\mathcal{L}_{\text{VRNF}}(\bm{\omega}, \tilde{\bm{s}}_{1:T}) = \sum_{t=1}^{T} \bigg\{  \frac{1}{L} \sum_{i=1}^{L} \log p(\bm{y}_{t} | \bm{x}_{t}^{(i)} (\tilde{\bm{s}}_t))\bigg\} \nonumber \\
 - KL \big( q(\bm{x}_{1:T} )~||~p(\bm{x}_{1:T} ) \big), 
\end{align}
where $L$ is the number of samples used for calibration, $\bm{x}_{k}^{(i)} (\tilde{\bm{s}}_k)$ is the $i$-th sample given the latent state $\tilde{\bm{s}}_k$, and $KL \big( \cdot )$ is the KL divergence term defined based on the priors in Section \ref{sec:prior_spec}. 

\subsection{VAE Priors for VRNF}
\label{sec:prior_spec}
Using the generative model for $\bm{x}_t$ in Equation \eqref{eqn:generative_model_x}, we consider the definition of two priors for the VRNF, as described briefly below. A full definition can be found in Appendix\footnote{URL for full paper with appendix: {\color{blue}\underline{{\url{https://arxiv.org/abs/1901.08096}}}}} \ref{apdx:kl_term}, which also includes derivations for the KL term used in $\mathcal{L}_{\text{VRNF}}(\bm{\omega}, \tilde{\bm{s}}_{1:T})$.

\textbf{Kalman Filter Prior (VRNF-KF) \quad} Considering a linear Gaussian state space form for Equations \eqref{eqn:generative_model_y} and \eqref{eqn:generative_model_x}, we can apply the Kalman filtering equations to obtained distributions for $\bm{x}_t$ at each time step (e.g. $p(\bm{x}_t | \bm{y}_{1:t}, \bm{u}_{1:t})$). This also lets us analytically define how the means and covariances of the belief state change with different sets of information -- aligning the VRNF's encoder stages with the filtering equations.

\textbf{Neural Network Prior (VRNF-NN) \quad} In the spirit of the DKF \cite{DeepKalmanFilters}, the analytical equations from the Kalman filter prior above can also be approximated using simple multilayer perceptrons. This would also allow belief state updates to accommodate non-linear states space dynamics, making it fa less restrictive prior model.

\subsection{Encouraging Decoupled Representations}
\textbf{Combined Encoder Training \quad} To improve representation learning, the RNF is trained in a ``multi-task'' fashion -- with each intermediate stage trained to encode latent states for output distributions. This is achieved by applying the same emissions decoder to all encoders during training as indicated in Figure \ref{fig:architecture}, with each encoder/decoder aligned with the Bayesian filtering steps described in Section \ref{sec:architecture}.  Encoders are then trained jointly using the combined loss function below:  
\begin{align}
&\mathcal{L}_{\text{combined}}(\bm{\omega}, \bm{y}_{1:T}, \bm{u}_{1:T}) \nonumber \\ &=\underbrace{\mathcal{L}(\bm{\omega}, \tilde{\bm{s}}_{1:T})}_{\text{Input Dynamics}} +  \overbrace{ \underbrace{\alpha_x\mathcal{L}(\bm{\omega}, \tilde{\bm{s}}^{'}_{1:T})}_{\text{Propagation}} + \underbrace{\alpha_y\mathcal{L}(\bm{\omega}, \bm{s}_{1:T})}_{\text{Error Correction}}}^{\text{Additional Regularisation Terms}}.
\end{align}
As such, the additional stages can be interpreted as regularisation terms for the VRNF or RNF loss functions -- which we weight by constants $\alpha_x$ and $\alpha_y$ to control the relative importance of the intermediate encoder representations. For our main experiments, we place equal importance on all encoders, i.e. $\alpha_x=\alpha_y =1$, to facilitate the subsequent separation of stages for multistep prediction -- with a full ablation analysis performed to assess the impact of various $\alpha$ settings during training. 

Furthermore, the error correction component $\phi_y(\cdot)$ can also be interpreted as a pure auto-encoding step for the latest observation, recovering distributions $p(\bm{y}_t | \bm{x}_t)$ based on filtered distributions of $p(\bm{x}_t | \bm{y}_{1:t}, \bm{u}_{1:t})$. Given that all stages share the same emissions decoder, this obliges the network to learn representations for $\bm{s}_t$ that are able to reconstruct the current observation when it is available.

\textbf{Introducing Artificial Missingness \quad} Next, to encourage the clean separation of encoder stages for generalisation to other tasks, we break dependencies between the encoders by introducing artificial missingness into the dataset -- randomly dropping out inputs and observations with a missingness rate $r$. As encoders are only applied where data is present (see Figure \ref{fig:missingness}), input dynamics and error correction encoders are hence randomly skipped over during training -- encouraging the encoder to perform regardless of which encoder stage preceded it. This also bears a resemblance to input dropout during training, which we apply to competing benchmarks to ensure comparability.

\section{Performance Evaluation} 
\label{sec:experiments}

\begin{table*}[ht]
\caption{Normalised MSEs For One-Step-Ahead Predictions}
\label{tab:one_step}
\centerline{
\begin{tabular}{@{}l|ll|ll|ll|l@{}}
\toprule
\textbf{}            & \textbf{DeepAR} & \textbf{DSSM} & \textbf{VRNN} & \textbf{DKF} & \textbf{VRNF-KF} & \textbf{VRNF-NN} & \textbf{RNF}   \\ \midrule
\textbf{Electricity} & 0.908           & 1.000         & 2.002         & 0.867        & 0.861            & 0.852            & \textbf{0.780*} \\
\textbf{Volatility}  & 3.956           & 1.000         & 0.991         & 0.982        & 1.914            & 1.284            & \textbf{0.976*} \\
\textbf{Quote}       & 0.998           & 1.000         & 3.733         & 1.001        & 1.000            & 1.001            & \textbf{0.997*} \\ \bottomrule
\end{tabular}}
\end{table*}

\begin{table*}[ht]
\caption{Coverage Probability Of One-Step-Ahead 90\% Prediction Interval}
\label{tab:pcip}
\centering
\begin{tabular}{@{}l|ll|ll|ll|l@{}}
\toprule
\textbf{}            & \textbf{DeepAR} & \textbf{DSSM} & \textbf{VRNN} & \textbf{DKF} & \textbf{VRNF-KF} & \textbf{VRNF-NN} & \textbf{RNF} \\ \midrule
\textbf{Electricity} & 0.966           & 0.964         & 0.981         & 0.965        & {\color{red} \textbf{0.320}}        & {\color{red} \textbf{0.271}}            & \textbf{0.961*}        \\
\textbf{Volatility}  & \textbf{0.997*}           & 0.999         & 1.000         & 1.000        & 1.000            & 1.000            & 1.000        \\
\textbf{Quote}       & 0.997           & 0.991         & {\color{red} \textbf{0.005}}         & 0.998        & \textbf{0.924          *} & 0.992            & 0.997        \\ \bottomrule
\end{tabular}
\end{table*}

\begin{table*}[ht]
\caption{Normalised MSEs For Multistep Predictions With Both Unknown and Known Inputs}
\label{tab:multistep}
\centering
\begin{tabular}{@{}l|ll|ll|ll|ll|l@{}}
\toprule
\textbf{Input Type}     & \textbf{Dataset}     & \textbf{$\tau=$} & \textbf{DeepAR} & \textbf{DSSM} & \textbf{VRNN} & \textbf{DKF} & \textbf{VRNF-KF} & \textbf{VRNF-NN} & \textbf{RNF}    \\ \midrule
\textbf{Unknown Inputs} & \textbf{Electricity} & 5                & 3.260           & 3.308         & 3.080         & 2.946        & 2.607            & 2.015            & \textbf{1.996*} \\
\textbf{}               & \textbf{}            & 10               & 4.559           & 4.771         & 4.533         & 4.419        & 5.467            & \textbf{3.506*}  & 3.587           \\
\textbf{}               & \textbf{}            & 20               & 6.555           & 6.827         & 6.620         & 6.524        & 9.817            & \textbf{5.449*}  & 6.098           \\ \cmidrule(l){2-10} 
\textbf{}               & \textbf{Volatility}  & 5                & 3.945           & 1.628         & 0.994         & 0.986        & 4.084            & 1.020            & \textbf{0.967*} \\
\textbf{}               & \textbf{}            & 10               & 3.960           & 1.639         & 0.994         & 0.985        & 4.140            & 1.017            & \textbf{0.967*} \\
\textbf{}               & \textbf{}            & 20               & 3.955           & 1.641         & 0.993         & 0.983        & 4.163            & 1.014            & \textbf{0.966*} \\ \cmidrule(l){2-10} 
\textbf{}               & \textbf{Quote}       & 5                & 1.000           & 1.000         & 1.001         & 1.000        & 1.002            & 0.999            & \textbf{0.998*} \\
\textbf{}               & \textbf{}            & 10               & 1.000           & 1.001         & 1.000         & 1.001        & 1.009            & 1.002            & \textbf{1.000*} \\
\textbf{}               & \textbf{}            & 20               & 1.000           & 1.001         & 1.003         & 1.001        & 1.488            & 1.003            & \textbf{1.000*} \\ \midrule
\textbf{Known Inputs}   & \textbf{Electricity} & 5                & 3.260           & 3.199         & 3.045         & 1.073        & 1.112            & 0.877            & \textbf{0.813*} \\
\textbf{}               & \textbf{}            & 10               & 4.559           & 4.382         & 4.470         & 1.008        & 1.180            & 0.882            & \textbf{0.831*} \\
\textbf{}               & \textbf{}            & 20               & 6.555           & 6.174         & 6.514         & 0.989        & 1.209            & 0.884            & \textbf{0.846*} \\ \cmidrule(l){2-10} 
\textbf{}               & \textbf{Volatility}  & 5                & 3.988           & 1.615         & 0.994         & 0.986        & 2.645            & 1.009            & \textbf{0.981*} \\
\textbf{}               & \textbf{}            & 10               & 3.992           & 1.620         & 0.994         & 0.985        & 2.652            & 1.009            & \textbf{0.981*} \\
\textbf{}               & \textbf{}            & 20               & 3.991           & 1.627         & 0.993         & 0.984        & 2.652            & 1.008            & \textbf{0.980*} \\ \cmidrule(l){2-10} 
\textbf{}               & \textbf{Quote}       & 5                & 1.000           & 1.000         & 0.998         & 1.000        & 1.001            & 1.000            & \textbf{0.997*} \\
\textbf{}               & \textbf{}            & 10               & 1.000           & 1.000         & 0.999         & 1.000        & 1.003            & 1.000            & \textbf{0.998*} \\
\textbf{}               & \textbf{}            & 20               & 1.000           & 1.000         & 1.003         & 1.000        & 1.009            & 1.000            & \textbf{0.999*} \\ \bottomrule
\end{tabular}
\end{table*}
%
\begin{table*}[ht]
\caption{Normalised MSEs for Ablation Studies}
\label{tab:ablation}
\centering
\begin{tabular}{@{}ll|llll|llll|llll@{}}
\toprule
\textbf{}               & \textbf{}       & \multicolumn{4}{c|}{\textbf{Electricity}}                             & \multicolumn{4}{c|}{\textbf{Volatility}}                              & \multicolumn{4}{c}{\textbf{Quote}}                                    \\
                        & $\tau=$         & 1               & 5               & 10              & 20              & 1               & 5               & 10              & 20              & 1               & 5               & 10              & 20              \\ \midrule
\textbf{Unknown Inputs} & \textbf{RNF}    & -               & \textbf{1.996*} & \textbf{3.587*} & \textbf{6.098*} & -               & \textbf{0.967*} & \textbf{0.967*} & \textbf{0.966*} & -               & \textbf{0.998*} & \textbf{1.000*} & \textbf{1.000*} \\
\textbf{}               & \textbf{RNF-NS} & -               & 2.801           & {\color{red}\textbf{13.409}}          & {\color{red}45.625}          & -               & 1.006           & 1.006           & 1.005           & -               & 1.137           & 1.294           & 1.260           \\
\textbf{}               & \textbf{RNF-IO} & -               & {\color{red}14.047 }         & {\color{red}14.803}          & {\color{red}15.414}          & -               & 1.377           & 1.458           & 1.494           & -               & 1.029           & 1.042           & 1.045           \\ \midrule
\textbf{Known Inputs}   & \textbf{RNF}    & 0.780           & 0.813           & \textbf{0.831*} & \textbf{0.846*} & \textbf{0.976*} & \textbf{0.981*} & \textbf{0.981*} & \textbf{0.980*} & \textbf{0.997*} & \textbf{0.997*} & \textbf{0.998*} & \textbf{0.999*} \\
                        & \textbf{RNF-NS} & 0.828           & 0.948           & 0.997           & 1.042           & 0.979           & 0.983           & 0.983           & 0.982           & 1.001           & 1.003           & 1.019           & 1.026           \\
                        & \textbf{RNF-IO} & \textbf{0.770*} & \textbf{0.809*} & 0.873           & 0.918           & 1.012           & 1.016           & 1.016           & 1.015           & 1.020           & 1.015           & 1.023           & 1.030           \\ \bottomrule
\end{tabular}
\end{table*}

\subsection{Time Series Datasets}
We conduct a series of tests on 3 real-world time series datasets to evaluate performance:
\begin{enumerate}
\item \textbf{Electricity:} The public UCI Individual Household Electric Power Consumption Data \cite{UCIPower}
\item \textbf{Volatility:} A 30-min realised variance \cite{RealizedVol} dataset for 30 different stock indices
\item \textbf{Quote:} A high-frequency market microstructure dataset containing Barclays Level-1 quote data from Thomson Reuters Tick History (TRTH)
\end{enumerate}
Details on input/output features and preprocessing are fully documented in Appendix \ref{apdx:datasets} for reference.

\subsection{Conduct of Experiment}
\textbf{Benchmarks:  \quad} We compare the VRNF-KF, VRNF-NN and standard RNF against a range of autoregressive and RVAE benchmarks -- including the DeepAR Model \cite{DeepAR}, Deep State Space Model (DSSM) \cite{DeepStateSpaceModels}, Variational RNN (VRNN) \cite{VRNN}, and Deep Kalman Filter (DKF) \cite{DeepKalmanFilters}.  

For multistep prediction, we consider two potential use cases for exogenous inputs: (i) when future inputs are unknown beforehand and imputed using their last observed values, and (ii) when inputs are known in advance and used as given. When models require observations of $\bm{y}_t$ as inputs, we recursively feed outputs from the network as inputs at the next time step. These tweaks allow the benchmarks to be used for multistep prediction without modifying network architectures. For the RNF, we consider the application of the propagation encoder alone for the former case, and a combination of the propagation and input dynamics encoder for the latter -- as detailed in Section \ref{sec:rnf_missing_data}.

\textbf{Metrics: \quad} To determine the accuracy of forecasts, we evaluate the mean-squared-error (MSE) for single-step and multistep predictions, normalising each using the MSE of the one-step-ahead forecast for the best autoregressive model (i.e. the DSSM). For multistep forecasts, we measure the average squared error up to the maximum prediction horizon ($\tau$). As observations are 1D continuous variables for all our datasets, we evaluate uncertainty estimates using the prediction interval coverage probability (PICP) of a $90\%$ prediction interval, defined as:
\begin{align}
\mathrm{PICP} = \frac{1}{T} \sum_{t=1}^T c_t, 
\end{align}
\begin{align}
c_t & =  
 \begin{cases} 
 	1, ~\text{if } \psi(0.05, t) < y_t < \psi(0.95, t)\\
 	0, ~ \text{otherwise}
 \end{cases}
\end{align}
where $\psi(0.05, t)$ is the $5^{th}$ percentile of samples from $N\left(f (\bm{x}_t), \bm{\Gamma} \right) $. 

\textbf{Training Details:  \quad} Please refer to Appendix \ref{apdx:training} for full details of network calibration.

\subsection{Results and Discussion}
On the whole, the standard RNF demonstrates the best overall performance -- improving MSEs in general for one-step-ahead and multistep prediction. From the one-step-ahead MSEs in Table \ref{tab:one_step}, the RNF improves forecasting accuracy by $19.6\%$ on average across all datasets and benchmarks. These results are also echoed for multistep predictions in Table \ref{tab:multistep}, with the RNF beating the majority of baselines for all horizons and datasets. The only exception is the slight out-performance of another RNF variant (the VRNF-NN) on the Electricity dataset with unknown inputs -- possibly due to the adoption of a suitable prior for this specific dataset -- with the standard RNF coming in a close second.  The PICP results of Table \ref{tab:pcip} also show that performance is achieved without sacrificing the quality of uncertainty estimates, with the RNF outputting similar uncertainty intervals compared to other deep generative and autoregressive models. On the whole, this demonstrates the benefits of the proposed training approach for the RNF, which encourages decoupled representations using regularisation terms and skip training.

To measure the benefits of the skip-training approach and proposed regularisation terms, we also perform a simple ablation study and train the RNF without the proposed components. Table \ref{tab:ablation} shows the normalised MSEs for the ablation studies, with one-step and multistep forecasts combined into the same table. Specifically, we test the RNF with no skip training in RNF-NS, and the RNF with only input dynamics outputs included in the loss function (i.e. $\alpha_x=\alpha_y=0$) in RNF-IO.  As inputs are always known for one-step-ahead predictions, normalised MSEs for $\tau=1$ are omitted for unknown inputs. In general, the inclusion of both skip training and regularisation terms improves forecasting performance, particularly in the case of longer-horizon predictions. We observe this from the MSE improvements for all but short-term ($\tau \in \{1,5\}$) predictions for known inputs, where the RNF-IO. However, the importance of both skip-training and regularisation can be seen from the large multistep MSEs of both the RNF-NS and RNF-IO on the Electricity dataset with unknown inputs-- which results from error propagation when the input dynamics encoder is removed.

As mentioned in Section \ref{sec:rnf_section}, the challenges of prior selection for VAE-based methods can be seen from the PICPs in Table \ref{tab:pcip} -- with small PICPs for VRNF models indicative of miscalibrated distributions in the Electricity data, and the poor MSEs and PICPs for the VRNN indicative of posterior collapse on the Quote data. However, this can also be beneficial when applied to appropriate datasets -- as seen from the closeness of the VRNF-KF's PICP to the expected $90\%$ on the Quote data. As such, the autoregressive form of standard RNF leads to more reliable performance from both a prediction accuracy and uncertainty perspective -- doing away with the need to define a prior for $\bm{x}_t$.

\section{Conclusions}
In this paper, we introduce a novel recurrent autoencoder architecture, which we call the Recurrent Neural Filter (RNF), to learn decoupled representations for the Bayesian filtering steps -- consisting of separate encoders for state propagation, input and error correction dynamics,  and a common decoder to model emission. Based on experiments with three real-world time series datasets, the direct benefits of the architecture can be seen from the improvements in one-step-ahead predictive performance, while maintaining comparable uncertainty estimates to benchmarks. Due to its modular structure and close alignment with Bayesian filtering steps, we also show the potential to generalise the RNF to similar predictive tasks -- as seen from improvements in multistep prediction using extracted state transition encoders.
%

\bibliography{rnf_bib}
\bibliographystyle{IEEEtran}

\vfill
\newpage
\appendix

\subsection{VRNF Priors and Derivation of KL Term}
\label{apdx:kl_term}
Defining a prior distribution for the VRNF starts with the specification of a model for the distribution of hidden state $\bm{x}_t$, conditioned on the amount of available information at each encoder to achieve alignment with the VRNF stages. Per the generative model of Equation \eqref{eqn:generative_model_x}, we model $\bm{x}_t$ as a multivariate normal distribution with a mean and covariance that varies with time and the information present at each encoder, based on the notation below:\\

\noindent \textbf{Propagation:}
\begin{equation}
    p(\bm{x}_t | \bm{y}_{1:t-1}, \bm{u}_{1:t-1}) \sim N(\bm{\tilde{\beta}}_t^{'}, \bm{\tilde{\nu}}_t^{'}),
\end{equation}
\noindent \textbf{Input Dynamics:}
\begin{equation}
    p(\bm{x}_t | \bm{y}_{1:t-1}, \bm{u}_{1:t}) \sim N(\bm{\tilde{\beta}}_t, \bm{\tilde{\nu}}_t),
\end{equation}
\noindent \textbf{Error Correction:}
\begin{equation}
    p(\bm{x}_t | \bm{y}_{1:t}, \bm{u}_{1:t}) \sim N(\bm{\beta}_t, \bm{\nu}_t).
\end{equation}

For the various priors defined in this section, we adopt the use of diagonal covariance matrices for the inputs, defined as:
\begin{equation}
\bm{\nu}_t = diag(\bm{\gamma}_t \odot \bm{\gamma}_t),
\end{equation}
where $\bm{\gamma}_t \in \mathbb{R}^J$ is a vector of standard deviation parameters.

This approximation helps to reduce the computational complexity associated with the matrix multiplications using full covariance matrices, and the $O(J^2 T)$ memory requirements from storing full covariances matrices for an RNN unrolled across $T$ timesteps.\\

\noindent \underline{\textbf{KL Divergence Term}} \\

Considering the application of the input dynamics encoder alone (i.e. $\alpha_x=\alpha_y=0$), the KL divergence between independent conditional multivariate Gaussians at each time step can be hence expressed analytically as:
\begin{align}
&KL_{\text{Input}} \big( q(\bm{x}_{1:T} )~||~p(\bm{x}_{1:T} ) \big) \\
&=  \mathbb{E}_{q(\bm{x}_{1:T} )} \bigg[  \log \frac{p(\bm{x}_1)}{ q(\bm{x}_1 | \bm{\tilde{s}}_1)} && \nonumber \\
&+ \sum_{t=2}^T  \log \frac{p(\bm{x}_t | \bm{x}_{t-1}, \bm{u}_{1:t}, \bm{y}_{1:t-1})}{q(\bm{x}_t | \bm{\tilde{s}}_t)}  \bigg] &&\\
&= \sum_{t=1}^T \sum_{j=1}^J \bigg\{ \log \frac{ \tilde{\gamma}_t(j)}{\sigma(j, \tilde{\bm{s}}_t)} && \nonumber \\
&+  \frac{\sigma(j, \tilde{\bm{s}}_t)^2 + (m(j,\tilde{\bm{s}}_t) - \tilde{\beta}_t(j) )^2}{2 \tilde{\gamma}_t(j)^2}  - \frac{1}{2} \bigg\},&&
\end{align}
where $m(j, \tilde{\bm{s}}_t), \sigma(j, \tilde{\bm{s}}_t)$ are $j$-th elements of $m(\tilde{\bm{s}}_t), \sigma(\tilde{\bm{s}}_t)$ as defined in Equations \eqref{eqn:vrnf_encoder_mean} and \eqref{eqn:vrnf_encoder_std} respectively. 

The KL divergence terms are defined similarly for the propagation and error correction encoders, using the means and standard deviations defined above.\\

\noindent \underline{\textbf{Kalman Prior (VRNF-KF)}}\\

The use Kalman filter relies on the definition of a linear Gaussian state space model, which we specify below:
\begin{align}
\bm{y}_t &= \bm{H} \bm{x}_t + \bm{e}_t,\\
\bm{x}_t &= \bm{A} \bm{x}_{t-1} + \bm{B} \bm{u}_t + \bm{\epsilon}_t,
\end{align} 
where $\bm{H}, \bm{A}, \bm{B}$ are constant matrices, and $\bm{e}_t \sim N(0, \bm{R}), \bm{\epsilon}_t \sim N(0, \bm{Q})$ are noise terms with constant noise covariances $\bm{R}$ and $\bm{Q}$.

\paragraph{Propagation} Assuming that inputs $\bm{u}_t$ is unknown at time $t$, predictive distributions can still be computed for the hidden state if we have a model for $\bm{u}_t$. In the simplest case, this can be a standard normal distribution, i.e. $\bm{u}_t \sim N(\bm{c}, \bm{D})$ -- where $\bm{c}$ is a constant mean vector and $\bm{D}$ a constant covariance matrix. Under this model, predictive distributions can be computed as below:
\begin{align}
\label{eqn:prop_mean}
\tilde{\bm{\beta}}^{'}_t  &= \bm{A}\bm{\beta}_{t-1} + \bm{B} \bm{c} \nonumber \\
&= \bm{A} \bm{\beta}_{t-1} + \bm{c^{'}}, \\
\tilde{\bm{\nu}}^{'}_t & = \bm{A} \bm{\nu}_{t-1}  \bm{A}^T + \bm{B} \bm{D} \bm{B}^T + \bm{Q} \nonumber\\ 
 & = \bm{A} \bm{\nu}_{t-1}   \bm{A}^T +  \bm{Q}^{'},
 \label{eqn:prop_var}
\end{align}
with $\bm{c^{'}}, \bm{Q}^{'}$ collapsing constant terms together into a single parameters.

\paragraph{Input Dynamics} When inputs are known, the forecasting equations take on a similar form:
\begin{align}
\tilde{\bm{\beta}}_t  &= \bm{A}\bm{\beta}_{t-1}  + \bm{B} \bm{u}_t \\
\tilde{\bm{\nu}}_t & = \bm{A} \bm{\nu}_{t-1}  \bm{A}^T + \bm{Q} 
\end{align}

Comparing this with the forecasting equations of the propagation step, we can express also the above as functions $\tilde{\bm{\beta}}^{'}_t$  and $\tilde{\bm{\nu}}^{'}_t$, i.e.: 
\begin{align}
\label{eqn:ip_cond_on_prop_means}
\tilde{\bm{\beta}}_t  &= \tilde{\bm{\beta}}^{'}_t + \bm{B} \bm{u}_t - \bm{c^{'}}, \\
\tilde{\bm{\nu}}_t & = \tilde{\bm{\nu}}_t^{'} - \bm{Q}^{'} + \bm{Q}. 
\label{eqn:ip_cond_on_prop_stds}
\end{align}

\paragraph{Error Correction} Upon receipt of a new observation, the Kalman filter computes a Kalman Gain $\bm{K}_t$, using it to correct the belief state as below:
\begin{align}
\bm{\beta}_t  & = \left(\bm{I} - \bm{K}_t \bm{H}\right) \tilde{\bm{\beta}}_t - \bm{K}_t \bm{H} \bm{y}_t, \\
\bm{\nu}_t    & = \left(\bm{I} - \bm{K}_t \bm{H}\right) \tilde{\bm{\nu}}_t,
\end{align}
where $\bm{I}$ is an identity matrix and $\bm{K}_t = \tilde{\bm{\nu}}_t \bm{H}^T \left( \bm{H} \tilde{\bm{\nu}}_t \bm{H}^T + \bm{R} \right)^{-1}$. \\

\noindent \textit{\textbf{Approximations for Efficiency}}\\

To avoid the complex memory and space requirements associated with full matrix computations, we make the following approximations in our Kalman Filter equations.

\paragraph{Constant Kalman Gain} Firstly, as noted in \cite{ConstantKalmanGain}, Kalman gain values in stable filters usually tend towards a steady state value after a initial transient period. We hence fix the Kalman gain at a constant value, and collapse constant coefficients in the error correction equations to give:
\begin{align}
\bm{\beta}_t  & =  \bm{K}^{'} \tilde{\bm{\beta}}_t - \bm{H}^{'} \bm{y}_t, \\
\bm{\nu}_t    & =  \bm{K}^{'} \tilde{\bm{\nu}}_t,
\end{align}
Where $\bm{K}^{'} = \left(\bm{I} - \bm{K} \bm{H}\right) $ and $\bm{H}^{'} = \bm{K}\bm{H}$.

\paragraph{Independent Hidden State Dimensions} Next, we assume that hidden state dimensions are independent of one another, which effectively diagonalising state related coefficients $\bm{A} = diag(\bm{a})$ and $\bm{Q}= diag(\bm{q})$.

\paragraph{Diagonalising  $\bm{Q}^{'}, \bm{K}^{'}$} Finally, to allow us to diagonal covariance matrices throughout our equations, we also diagonalise $\bm{Q}^{'} = diag(\bm{q}^{'})$ and $\bm{K}^{'} = diag(\bm{k}^{'})$.\\

\noindent \textit{\textbf{Prior Definition}}\\

Using the above definitions and approximations, the Kalman filter prior can hence be expressed in vector form using the equations below:\\

\noindent\textbf{Propagation:}
\begin{align}
\tilde{\bm{\beta}}^{'}_t  &= \bm{a} \odot m(\bm{s}_{t-1}) + \bm{c^{'}}, \\
\tilde{\bm{\nu}}^{'}_t & = \bm{a} \odot V(\bm{s}_{t-1})  \odot \bm{a} +  \bm{q}^{'}.
\end{align}

\noindent \textbf{Input Dynamics:}
\begin{align}
\tilde{\bm{\beta}}t  &= \bm{a} \odot m(\bm{s}_{t-1}) + \bm{B} \bm{u}_t, \\
\tilde{\bm{\nu}}_t & = \bm{a} \odot V(\bm{s}_{t-1})  \odot \bm{a} +  \bm{q}.
\end{align}

\noindent \textbf{Error Correction:}
\begin{align}
\bm{\beta}_t  & =  \bm{k}^{'} \odot m(\bm{\tilde{s}}_{t})- \bm{H}^{'} \bm{y}_t, \\
\bm{\nu}_t    & =  \bm{k}^{'} \odot V(\bm{\tilde{s}}_{t}).
\end{align}

All constant standard deviation are implemented as coefficients wrapped in a softmax layer (e.g. $\bm{a} = \text{softplus}(\phi))$) to prevent the optimiser from converging on invalid negative numbers.

In addition, we note that the form input dynamics prior is not conditioned on the propagation encoder outputs, although we could in theory express it terms of its statistics (i.e. Equations \eqref{eqn:ip_cond_on_prop_means} and \eqref{eqn:ip_cond_on_prop_stds}). This to avoid converging on negative values for variances, which can be obtained from the subtraction of positive constant $\bm{Q^{'}}$, although we revisit this in this form in the next section.\\

\noindent \underline{\textbf{Neural Network Prior (VRNF-NN)}}\\

Despite the convenient tractable from of the Kalman filtering equations, this relies on the use of linear state space assumptions which might not be suitable for complex datasets. As such, we also consider the use of multilayer perceptrons $\left( \text{MLP}(\cdot) \right)$ to approximate the equations described in the previous section, conditioning it on the previous active encoder stage, i.e.:

\textbf{Propagation:}
\begin{align}
\tilde{\bm{\beta}}^{'}_t  &= \text{MLP}_{\tilde{\bm{\beta}}^{'}} (m(\bm{s}_{t-1}))\\
\tilde{\bm{\nu}}^{'}_t & =  \text{MLP}_{\tilde{\bm{\nu}}^{'}} (V(\bm{s}_{t-1}))
\end{align}

\textbf{Input Dynamics:}
\begin{align}
\tilde{\bm{\beta}}_t  &= \text{MLP}_{\tilde{\bm{\beta}}} \left( m(\tilde{\bm{s}}_{t}), \bm{u}_t \right)\\
\tilde{\bm{\nu}}_t & =  \text{MLP}_{\tilde{\bm{\nu}}} \left( V(\tilde{\bm{s}}_{t}), \bm{u}_t \right)
\end{align}

\textbf{Error Correction:}
\begin{align}
\bm{\beta}_t  &= \text{MLP}_{\bm{\beta}} \left( m(\bm{s}_{t}), \bm{y}_t \right)\\
\bm{\nu}_t & =  \text{MLP}_{\bm{\nu}} \left( V(\bm{s}_{t}), \bm{y}_t \right)
\end{align}

Similar to the state transition functions used in \cite{DeepKalmanFilters}, this can be interpreted as using MLPs to approximate the true Kalman filter functions for linear datasets, while also permitting the learning of more sophisticated non-linear models. All MLPs defined here use an ELU activation function for their hidden layer, fixing the hidden state size to be $J$. Furthermore, we use linear output layers for $\bm{\beta}$ MLPs, while passing that of  $\bm{\nu}$ MLPs through a softplus activation function to maintain positivity.

\subsection{Training Procedure for RNF}
\label{apdx:training}
\paragraph{Training Details} During network calibration, trajectories were partitioned into segments of 50 time steps each -- which were randomly combined to form minibatches during training. Also, networks were trained for up to a maximum of 100 epochs or convergence. For the electricity and volatility datasets, 50 iterations of random search were performed, using the grid found in Table \ref{tab:hyperparam_ranges}. 20 iterations of random search were used for the quote dataset, as the significantly larger dataset led to longer training times for a given set of hyperparameters. 

\begin{table}[htbp]
\caption{Random Search Grid for Hyperparameter Optimisation}
\label{tab:hyperparam_ranges}
\centering
\begin{tabular}{@{}ll@{}}
\toprule
\textbf{}               & \textbf{Hyperparameter Ranges}    \\ \midrule
\textbf{Dropout Rate}   & 0.0, 0.1, 0.2, 0.3, 0.4, 0.5             \\
\textbf{State Size}     & 5, 10, 25, 50, 100, 150             \\
\textbf{Minibatch Size} & 256, 512, 1024                      \\
\textbf{Learning Rate}  & 0.0001, 0.001, 0.01, 0.1, 1.0       \\
\textbf{Max Gradient Norm}       & 0.0001, 0.001, 0.01, 0.1, 1.0, 10.0 \\ 
\textbf{Missing Rate}       & 0.25, 0.5, 0.75  \\ 
\bottomrule
\end{tabular}
\end{table}

\paragraph{State Sizes} To ensure that consistency across all models used, we constrain both the memory state of the RNN and the latent variable modelled to have the same dimensionality -- i.e. $J = \mathrm{dim}(\bm{s}_t) = \mathrm{dim}(\bm{x}_t)$ for the RNF. The exception is the DSSM, as the full covariance matrix of the Kalman filter would result in a prohibitive $J^2$ memory requirement if left unchecked. As such, we use the constraint where both the RNN and the Kalman filter to have the same memory capacity for the DSSM -- i.e.  $J = \mathrm{dim}(\bm{s}_t) = \mathrm{dim}(\bm{x}_t) + \mathrm{dim}(\bm{x}_t)^2$.

\paragraph{Dropout Application} Across all benchmarks, dropout was applied only onto the memory state of the RNNs ($\bm{h}_t$) in the standard fashion and not to latent states $\bm{x}_t$. For the LSTM, DeepAR Model and RNF, this corresponds to applying dropout masks to the outputs and internal states of the network. For the VRNN, DKF and DSSM, we apply dropout only to the inputs of the network -- in line with \cite{StructuredInferenceNetworks} to maintain comparability to the encoder skipping in the VRNFs.

\paragraph{Artificial Missingness} Encoder skipping is restricted to only the VRNFs and standard RNF, controlled by the missing rates defined above.

\paragraph{Sample Generation} At prediction time, latent states for the VRNN, DKF and VRNFs are sampled as per the standard VAE case -- using $L=1$ during training, $L=30$ for our validation error and $L=100$ for at test time. Predictions from the DeepAR Model, DSSM and standard RNF, however, were obtained directly from the mean estimates, with that of the DSSM computed analytically using the Kalman filtering equations. While this differs slightly from the original paper \cite{DeepStateSpaceModels}, it also leads to improvements in the performance DSSM by avoiding sampling errors.

\subsection{Description of Datasets}
\label{apdx:datasets}
For the experiments in Section \ref{sec:experiments}, we focus on the use of 3 real-world time series datasets, each containing over a million time steps per dataset. These use-cases help us evaluate performance for scenarios in which real-time predictions with RNNs are most beneficial -- i.e. when the underlying dynamics is highly non-linear and trajectories are long.\\

\noindent \underline{\textbf{Summary}}\\

\textbf{Electricity: \quad} The UCI Individual Household Electric Power Consumption Dataset \cite{UCIPower} is a time series of 7 different power consumption metrics measured at 1-min intervals for a single household between December 2006 and November 2010 -- coming to a total of 2,075,259 time steps over 4 years. In our experiments, we treated active power as the main observation of interest, taking the remainder to be exogenous inputs into the RNNs.

\textbf{Intraday Volatility: \quad} We compute 30-min realised variances \cite{RealizedVol} for a universe of 30 different stock indices -- derived using 1-min index returns subsampled from Thomson Reuters Tick History Level 1 (TRTH L1) quote data. On the whole, the entire dataset contains 1,706,709 measurements across all indices, with each trajectory spanning 17 years on average. Given the strong evidence for the intraday periodicity of returns volatility \cite{IntradayRVSeasonality}, we also include the time-of-day as an additional exogenous input. 

\textbf{High-Frequency Stock Quotes: \quad} This dataset consists of extracted features from TRTH L1 stock quote data for Barclays (BARC.L) -- specifically forecasting microprice returns \cite{microprice} using volume imbalance as an input predictor -- comprising a total of 29,321,946 time steps between 03 January 2017 to 29 December 2017. From \cite{volumeimbalance}, volume imbalance in the limit order book is a good predictor of the direction (sign) of the next liquidity taking order, and the price changes immediately after the arrival of a liquidity-taking order. \\

\noindent \underline{\textbf{Electricity}} \\

\paragraph{Data Processing} The full trajectory was segmented into 3 portions, with the earliest 60\% of measurements for training, the next 20\% as a validation set, and the final 20\% as an independent test set -- with results reported in Section \ref{sec:experiments}. All data sets were normalised to have zero mean and unit standard deviations, with normalising constants computed using the training set alone.

\paragraph{Summary Statistics} A list of summary statistics can be seen in Table \ref{tab:stats_power}.
\begin{table}[ht]
\caption{Summary Statistics for Electricity Dataset}
\label{tab:stats_power}
\centering
\begin{tabular}{lllll}
\hline
\textbf{}               & \textbf{Mean} & \textbf{S.D.} & \textbf{Min} & \textbf{Max} \\ \hline
\textbf{Active Power*}   & 1.11          & 1.12          & 0.08         & 11.12        \\
\textbf{Reactive Power} & 0.12          & 0.11          & 0.00         & 1.39         \\
\textbf{Intensity}      & 4.73          & 4.70          & 0.20         & 48.40        \\
\textbf{Voltage}        & 240.32        & 3.33          & 223.49       & 252.72       \\
\textbf{Sub Metering 1} & 1.17          & 6.31          & 0.00         & 82.00        \\
\textbf{Sub Metering 2} & 1.42          & 6.26          & 0.00         & 78.00        \\
\textbf{Sub Metering 3} & 6.04          & 8.27          & 0.00         & 31.00        \\ \hline
\end{tabular}
\end{table}

\noindent \underline{\textbf{Intraday Volatility}}\\

\paragraph{Data Processing} From the 1-min index returns, realised variances were computed as:
\begin{align}
r_k &= \ln p_k - \ln p_{k-1} \nonumber\\ 
y(t, 30) &= \sum_{k=t-30}^t r_k ^2,
\end{align}
where $r_k$ is the 1-min index return at time $k$, $\ln p_k$ is the log price at $k$, and $y(t, 30)$ is the 30-min realised variance at time $t$.

Before computation, the data was cleaned by only considering prices during exchange hours to avoid spurious jumps. In addition, realised variances greater than 10 times the 200-step rolling standard deviation were removed and replaced by its previous value -- so as to reduce the impact of outliers.

For the experiments in Section \ref{sec:experiments}, data across all stock indices were grouped together for training and testing -- using data prior to 2014 for training, data between 2014-2016 for validation and data from 2016 to 4 July 2018 for independent testing. Min-max normalisation was applied to the datasets, with time normalised by the maximum trading window of each exchange and realised variances by the max and min values of the training dataset.

\paragraph{Stock Index Identifiers (RICs):} AEX, AORD, BFX, BSESN, BVLG, BVSP, DJI, FCHI, FTMIB, FTSE, GDAXI, GSPTSE, HSI, IBEX, IXIC, KS11, KSE, MXX, N225, NSEI, OMXC20, OMXHPI, OMXSPI, OSEAX, RUT, SMSI, SPX, SSEC, SSMI, STOXX50E

\paragraph{Summary Statistics} A table of summary statistics can be found in Table \ref{tab:stats_vol} and give an indication of the general ranges of trajectories. 

\begin{table}[h]
\caption{Summary Statistics for Volatility Dataset}
\label{tab:stats_vol}
\centering
\begin{tabular}{@{}lllll@{}}
\toprule
\textbf{}                  & \textbf{Mean} & \textbf{S. D.} & \textbf{Min} & \textbf{Max} \\ \midrule
\textbf{Realised Variance*} & 0.0007        & 0.0017         & 0.0000       & 0.1013       \\
\textbf{Normalised Time}   & 0.43          & 0.27           & 0.00         & 0.97 \\ \bottomrule
\end{tabular}
\end{table}

\noindent \underline{\textbf{High-Frequency Stock Quotes}}\\

\paragraph{Input/Output Definitions} Microprice returns $y_t$ are defined as:
\begin{align*}
p_t &= \frac{V_a(t) p_b(t) + V_b(t) p _a(t)}{ V_a(t) + V_b(t)} \\
y_t & = \frac{p_t - p_{t-1} }{p_{t-1}}
\end{align*}
Where $V_b(t)$ and $V_a(t)$ are the bid and ask volumes at time $t$ respectively,  $p _b(t)$ and $p _a(t)$ are the bid/ask prices, and $p_t$ the microprice.

Volume imbalance $I_t$ is then defined as:
\begin{align*}
I_t = \frac{V_b(t) - V_a(t)}{ V_b(t) + V_a(t)} 
\end{align*}

\paragraph{Data Processing} From the raw Level 1 (best bid and ask prices and volumes) data from TRTH, we isolate measurements between 08.30 to 16.00 UK time, avoiding the effects of opening and closing auctions in our forecasts. Furthermore, microprice returns were also normalised using an exponentially weighting moving standard deviation with a half-life of 10,000 steps. We note that volume imbalance by definition is restricted to be $I_t \in [-1, 1]$, and hence does not require additional normalisation. Finally, the data was partitioned with training data from January to June, validation data from June to September, and the remainder for independent testing.

\paragraph{Summary Statistics}  Basic statistics can be found in Table \ref{tab:quote_stats}, and give an indication of the range of different variables.

\begin{table}[h]
\caption{Summary Statistics for Quote Dataset}
\label{tab:quote_stats}
\centering
\begin{tabular}{@{}lllll@{}}
\toprule
\textbf{}                   & \textbf{Mean} & \textbf{S. D.} & \textbf{Min} & \textbf{Max} \\ \midrule
\textbf{Normalised Returns*} & 0.00          & 0.80           & -117.72      & 117.13       \\
\textbf{Volume Imbalance}    & 0.02          & 0.48           & -1.00        & 1.00         \\ \bottomrule
\end{tabular}
\end{table}

\end{document}